\documentclass[11pt]{article}
\usepackage{coling2020}
\usepackage{multirow}
\usepackage{rotating}
\usepackage{times}
\usepackage{graphicx}
\usepackage{subcaption}
\usepackage{adjustbox}
\usepackage{footmisc}
\usepackage{url}
\usepackage{latexsym}
\usepackage{array}
\newcolumntype{M}[1]{>{\centering\arraybackslash}m{#1}}
\usepackage{xcolor}
\pdfoutput=1

\colingfinalcopy 

\title{Conversational Machine Comprehension: a Literature Review}

\author{Somil Gupta \\
  CICS, UMass Amherst \\
  \texttt{somilgupta@umass.edu}
  \And
  Bhanu Pratap Singh Rawat \\
  CICS, UMass Amherst \\
  \texttt{brawat@umass.edu}
  \And
   Hong Yu\\
  CICS, UMass Lowell \\
  \texttt{Hong\_Yu@uml.edu}
  }

\date{Oct 30, 2020}

\begin{document}
\maketitle
\begin{abstract}
Conversational Machine Comprehension (CMC), a research track in conversational AI, expects the machine to understand an open-domain natural language text and thereafter engage in a multi-turn conversation to answer questions related to the text. 
While most of the research in Machine Reading Comprehension (MRC) revolves around single-turn question answering (QA), multi-turn CMC has recently gained prominence, thanks to the advancement in natural language understanding via neural language models such as BERT and the introduction of large-scale conversational datasets such as CoQA and QuAC. The rise in interest has, however, led to a flurry of concurrent publications, each with a different yet structurally similar modeling approach and an inconsistent view of the surrounding literature. With the volume of model submissions to conversational datasets increasing every year, there exists a need to consolidate the scattered knowledge in this domain to streamline future research. This literature review attempts at providing a holistic overview of CMC with an emphasis on the common trends across recently published models, specifically in their approach to tackling conversational history. The review synthesizes a generic framework for CMC models while highlighting the differences in recent approaches and intends to serve as a compendium of CMC for future researchers.
\end{abstract}


\blfootnote{
    
    \hspace{-0.65cm}  
    This work is licensed under a Creative Commons 
    Attribution 4.0 International License.
    License details:
    \url{http://creativecommons.org/licenses/by/4.0/}.
}
\section{Introduction}
\label{introduction}
\par Developing open-domain, intelligent dialog systems that can satisfactorily interact like humans, perform complex tasks and/or answer on a range of topics has been one of the most ambitious and difficult goals in Artificial Intelligence (AI). The study of such systems, called \textbf{Conversational AI} (ConvAI), is at the confluence of Natural language Processing (NLP), Information Retrieval (IR), and Machine Learning (ML), attracting significant research from both academia and industry. The recent developments in Deep Learning (DL) \cite{nn-boost-1,nn-boost-2} and Reinforcement Learning (RL) \cite{rl-boost-1,rl-boost-2} have further boosted research in the domain, making it one of the most sought after research topics in AI. 


\par Based on the nature of problems, a ConvAI system is expected to solve three major research problems \cite{Survey-2010s}. \textit{Question Answering} (QA) involves providing answers to user queries through conversation, using the knowledge drawn from various data sources like a snippet from a text, a collection of web documents, or an entire knowledge base. \textit{Task completion} expects the conversational agent to accomplish task/s for the user, using the information acquired through conversation. Finally, \textit{Social Chat} makes the agent emulate humans and converse seamlessly and appropriately with users, as in the Turing test \cite{saygin2000turing}. Each of these fields has its own set of challenges to tackle.

\par Challenges in QA can vary depending on the source of knowledge, the answer extraction strategy employed, and the domain of the question. \textbf{Machine Reading Comprehension (MRC)} is one such challenge in QA, that requires the conversational QA (ConvQA) agent to understand a given open-domain text and thereafter answer question/s in conversation about it. These questions are often not paraphrased and may co-reference previous queries. The required answer may be a span of the given text or free-form. When the machine comprehension dialog involves multiple co-referenced questions such that a latter question may be a logical successor of the former, the challenge is termed as \textbf{Conversational Machine Comprehension} (CMC). A lot of research in MRC revolves around single-turn QA, but multi-turn CMC also holds major relevance because humans seek information conversationally by asking follow-up questions for additional information based on what they have already learned. Still, the inherent complexity involved in dealing with text comprehension and reasoning over dialogs and context had kept CMC as a far-fetched goal. However, the recent success in achieving at-par-with-human performance on single-turn MRC models \cite{squad2.0} due to the advancement in natural language understanding and modeling \cite{bert,albert,RoBERTa}, and the introduction of large-scale conversational datasets CoQA \cite{Coqa} and QuAC \cite{quac} have made information-seeking dialogs possible.

\par As a consequence, CMC has seen a significant surge in research in recent years. In less than 2 years since the introduction of these datasets, there have been 40 submissions\footnote{\label{current}Recorded as of July 1, 2020. Please note that many of these submissions are either ensemble versions of single models, or hyper-parameter variants of their pre-published models or are simply unpublished. Therefore, unique published models' count is 13 in CoQA and 7 in QuAC.} to CoQA leaderboard\footnote{\label{coqa_l}\url{https://stanfordnlp.github.io/coqa/}} and 22 submissions\footref{current} to QuAC leaderboard\footnote{\label{quac_l}\url{http://quac.ai/}}. Many of these models are unpublished, indicating active ongoing research on these datasets. Besides, the current state-of-the-art in QuAC lags behind human performance F1 benchmark by a margin of 6.7\footref{current}, suggesting significant scope for improvement. Almost simultaneously, there have been breakthroughs in NLP \cite{bert,gpt,RoBERTa} which the researchers have tried to leverage in their upcoming models \cite{ham,flowDelta,GraphFlow}. 
Since many models are being published concurrently, there have been inconsistencies and/or overlap in their methodologies and research directions. This makes it difficult to compare different approaches against each other and weigh their pros and cons. 
This prevailing scenario has blurred the bigger picture and made it difficult for researchers to attend to novel research in this field.
Moreover, there is no singular summarized view on CMC models, except the individual literature studies of these publications which can be highly localized and inconsistent with the global view. Thus, the current scenario motivates the need for organizing the scattered knowledge across these publications into a consolidated overview, so that future research in this field can be streamlined.

\par This literature review, therefore, provides a bird-eye overview of Conversational Machine Comprehension. We commence with an introduction to CMC, acquainting the reader with the challenges that make CMC unique, and the large-scale conversational datasets that spurred research in this field. \textit{To develop a general understanding of the CMC approaches, we shift the focus from comprehending individual models to observing the common trends that mark these models,  synthesizing a generic framework for a CMC model in the process.} We finally end our review with a discussion on the current trends and suggest advancements in the future.

\section{Related Work}
\par There have been several published literature reviews on MRC in recent years. \newcite{Survey-2010s} provides an extensive review of Conversational AI with a detailed account of the neural approaches being employed in each of its dialog systems (QA, Task completion, and social chat). It briefly discusses the problem of CMC and its datasets but does not comment upon the recent advancements and prevalent approaches in this domain. \newcite{Survey-MRC} provides a summary of all the recent single-turn MRC  datasets and approaches, however, it briefly discusses CoQA but does not touch upon any approaches for CMC. \newcite{2019-Survey} summarizes the classic models of single-turn MRC with a focus on deriving a common architecture and suggesting improvements based on the analysis. CMC is mentioned as an emerging research direction in this survey. The latest review by \newcite{2020-Survey} provides an overview of MRC along with the statistical analysis of datasets and the various problems in this domain. It mentions CMC as an MRC challenge but does not provide any further details.


\par This review differs from its predecessors as it focuses primarily on Conversational (multi-turn) Machine Comprehension which has not been detailed in the previous literature. CMC has its own set of challenges and an active research community around it. This calls for considering CMC as a separate research direction from single-turn MRC and review its rapid developments in terms of its general trends. 


\section{What is Conversational Machine Comprehension?}
\label{CMC}
\par The task of CMC is defined as: \textit{Given a passage $P$, the conversation history in the form of question-answer pairs
\{$Q_1,A_1,Q_2,A_2,...,Q_{i-1},A_{i-1}\}$ and a question $Q_i$, the model needs to predict the answer $A_i$}. The answer $A_i$ can either be a text span $(s_i, e_i)$ \cite{quac} or a free-form text $\{a_{i,1},a_{i,2},..., a_{i,j}\}$ with evidence $R_i$ \cite{Coqa}. Single-turn MRC models cannot directly cater to CMC, as the latter is much more challenging to address. The major challenges being:
\begin{itemize}
\itemsep0em
    \item The encoding module needs to encode not only $P$ and $A_i$ but also the conversational history. 
    \item General observation about information-seeking dialog in humans suggests that the starting dialog-turns tend to focus on the beginning chunks of the passage and shift focus to the later chunks as the conversation progresses \cite{quac}. The model is thus expected to capture these focal shifts during a conversation and reason pragmatically, instead of only matching lexically or via paraphrasing.
    \item Multi-turn conversations are generally incremental and co-referential. These conversational dialogs are either \textit{drilling down} (the current question is a request for more information about the topic), \textit{shifting topic} (the current question is not immediately relevant to something previously discussed), \textit{returning topic} (the current question is asking about a topic again after it had previously been shifted away from), \textit{clarification} of topic, or \textit{definition }of an entity \cite{dataset_comparison}. The model should, therefore, be able to take context from history which may or may not be immediate. 
\end{itemize} 

\section{Multi-Turn Conversational Datasets}
\par The surge in CMC research is credited to the emergence of large-scale multi-turn conversational datasets: CoQA \cite{Coqa} and QuAC \cite{quac}. 
\label{datasets}

\subsection{CoQA}
\par Conversational QA (CoQA) dataset consists of 126k questions sourced from 8k conversations. 
\begin{itemize}
\itemsep0em
    \item \textit{\textbf{Dataset preparation}}: Conversations are prepared over passages collected across 7 different domains, each with its source dataset, such as news articles derived from CNN \cite{cnn_daily}. Amongst the 7 domains, two are used for out-of-domain evaluation (only for evaluation, not training), while the other five aid in-domain evaluation (both training and evaluation). The dialog is prepared in a two annotator setting with one questioning and another answering, both referring to the entire context. 
    \item \textit{ \textbf{Questions:}} Questions are factoid but require sufficient co-referencing and pragmatic reasoning \cite{pragmaticReasoning}.
    \item \textit{\textbf{Answers}}: Answers are free-form, with their corresponding rationale highlighted in the passage. However, \newcite{dataset_comparison} identified that the answers are slightly modified versions of the rationale, and therefore optimizing an extractive model to predict the answer span with maximum F1 overlap to the gold answer can achieve up to 97.8 F1. 
    \item \textit{\textbf{Dialog features}}: The dialogs mostly involve drilling-down for details (about 60\% of all questions) but lack other dialog features like topic-shift, clarification, or definition. 
    \item \textit{\textbf{Evaluation}}: Macro-average F1 score of word overlap is used as an evaluation metric and is computed separately for in-domain and out-of-domain. 
\end{itemize}

\subsection{QuAC}
    \par Question Answering in Context (QuAC) contains 100K questions obtained from 14K information-seeking dialogs. 
\begin{itemize}
\itemsep0em
    \item \textit{\textbf{Dataset preparation}}: Dialogs are prepared over sections from Wikipedia articles about people from different genres such as culture and wildlife. The dataset is prepared using an asymmetric setting, with a student exposed only to the title of the article and a summary while the teacher is exposed to the entire section of the article on which the dialog is to be based. The student, therefore, tries to seek information about the hidden questions based on the limited information it gets from the dialog, and the teacher answers by providing short excerpts from the section (or `No Answer' if not possible).
    \item \textit{\textbf{Questions}}: Questions are descriptive, highly-contextual, and open-ended due to the asymmetric nature of the dataset that prevents paraphrasing. They require sufficient co-referencing and pragmatic reasoning. 
    \item \textit{\textbf{Dialog features}}: Besides drilling down, dialogs switch to new topics more frequently than CoQA. The dataset though lacks definition or clarification dialogs. 
    \item \textit{\textbf{Answers}}: Answers are extractive and can be either Yes/No or `No Answer'. Besides extractive span, the response also includes additional signals called dialog acts like continuation (follow up, maybe follow up, or don’t follow up) and affirmation (yes, no, or neither), which provides additional useful dialog flow information to train on, as used by \newcite{ham} and \newcite{Roberta+KD}. Further, an analysis of the answer token lengths in Table \ref{tab:comparison_quac_coqa} shows that QuAC answers are longer, which can be attributed to its asymmetric nature thereby motivating the seeker to ask open-ended questions to gauge hidden text.
    \item \textit{\textbf{Evaluation}}: Besides the macro-averaged F1 score on the entire set, QuAC also evaluates Human Equivalence Score (HEQ) to judge system performance relative to an average human, by finding the percentage of instances for which the system’s F1 matches or exceeds human F1. HEQ-Q and HEQ-D are thus HEQ scores with the instances as questions and dialogs respectively.
\end{itemize}
General dataset characteristics and an example from each of the datasets are provided in Appendix \ref{appdx:coqa_quac_example}.

\section{Generic Framework of a CMC Model}
\label{components}
\par
\begin{figure}[h!]
    \centering
    \makebox[\textwidth]{\includegraphics[width=15cm]{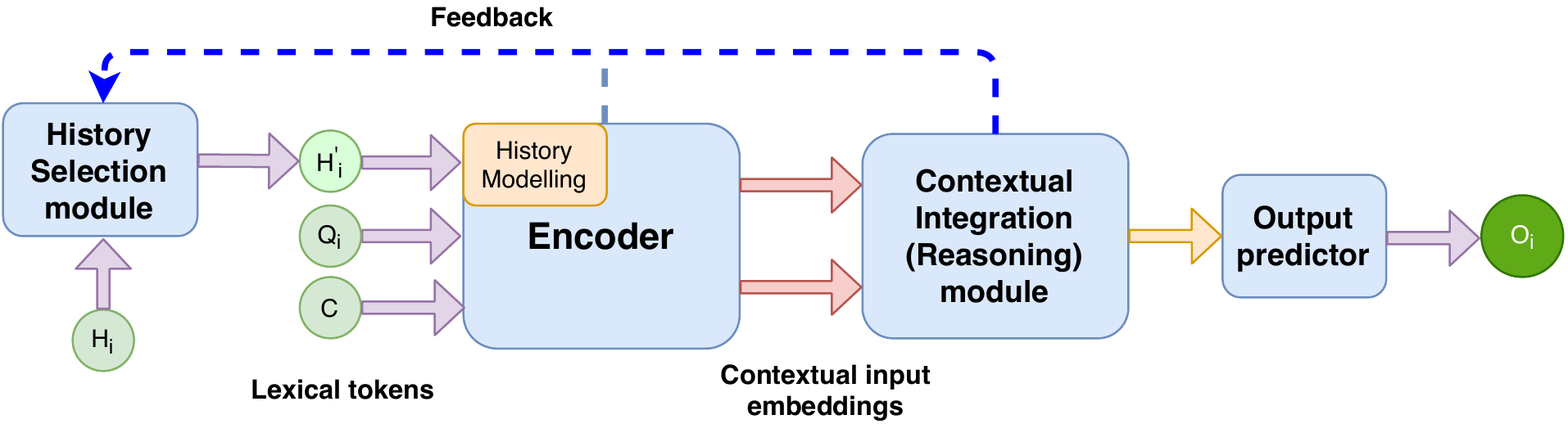}}
    \caption{Generic framework of a CMC model. A typical CMC model would consist of (1) History selection module, that selects a subset $H'_i$ of conversational history $H_i$ relevant to the current question $Q_i$; (2) Encoder, that encodes the lexical tokens of context $C$, $Q_i$ and $H'_i$ into input embeddings for contextual integration layer; (3) Reasoning module, that performs contextual integration of input embeddings into contextualized embeddings; and finally, (4) Output predictor, that predicts the output set $O_i$ based on contextualized embeddings.}
    \label{fig:cmc_model}
\end{figure}

 \newcite{Survey-2010s} defined the steps for performing reading comprehension in a typical neural MRC model as (1) \textit{encoding} the questions and context into a set of embeddings in a neural space;  (2) \textit{reasoning} in the neural space to identify the answer vector and (3) \textit{decoding} the answer vector into a natural language output. \newcite{flowQA} adapted these steps in CMC by adding conversational history modeling. \newcite{HAS} proposed a ConvQA model with separate modules for history selection and modeling. Based on these prior works, we synthesize a generic framework for a CMC model. A typical CMC model is provided with context $C$, current question $Q_i$ and the conversation history $H_i = [\{Q_{k},A_{k}\}]^{i-1}_{k=1}$, and needs to generate an output set $O_i$. The CMC framework is provided in Fig.~\ref{fig:cmc_model}. There are \textit{four} major components of the framework, based on their contribution to the overall CMC flow. 

\begin{enumerate}
\itemsep0em
    \item \textbf{History Selection module: } With complicated dialog behaviors like topic shift or topic return \cite{dataset_comparison}, simply selecting immediate turns may not work well. A history selection module, therefore, chooses a subset $H'_i$ of the history turns $H_i$ based on a policy (dynamic or static) that is expected to be more helpful than the others. If the history selection module is based on a dynamic learned policy (e.g. \newcite{ham}), then feedback from the other modules can guide its update.

    \item \textbf{Encoder:} The lexical tokens of the context passage $C$, selected conversational turns  $H'_i$,  and the current question  $Q_i$ need to be transformed into input embeddings for the reasoning module. Encoder facilitates this transition. The encoder steps may vary with every approach and reasoning inputs, at a high level, encoding involves transformation and combination of context-independent word embeddings called \textit{lexical embeddings} such as GloVE \cite{glove}, \textit{intra-sequence contextual embeddings} e.g. ELMo \cite{Elmo}, BERT \cite{bert} or RNN, \textit{question-aware embeddings}, and \textit{additional feature embeddings} like POS tags \cite{sdnet}, history embedding \cite{HAS} or conversation count. Conversational history $H'_i$ is generally integrated with this module into any or all of the contextual input embeddings. This process is called \textit{History modeling} and is the most significant aspect of a CMC encoder.   

    \item \textbf{Contextual Integration layer:} Contextual information accumulated in the passage, query, and/or history embeddings individually must be fused to generate query-aware and/or history-aware contextualized output embeddings. This process may involve a single layer (\textit{single-step reasoning}) or repetition across multiple layers (\textit{multi-step reasoning}). Input for this module generally consists of two (or more) sequence sets for every history turn, or aggregated across all turns, which are then fused in each layer and often inter-weaved \cite{fusionNet} with attention.

    \item \textbf{Output Predictor:} The model output may be in the form of a text span, signals like dialog acts \cite{quac} or a free-form (abstractive) answer \cite{Coqa}. Contextual embeddings generated by the reasoning module have all the latent information about the question, context passage, and conversational history. To get the token-level output, a fully-connected network followed by a softmax layer is generally used for per-token probability (abstractive) or start/end probability (extractive). Besides, a linear neural network may be used to find the aggregated result of the sequence. 
    
\end{enumerate}

\section{Common Trends across CMC models}
\par Instead of describing each CMC model separately, we categorized them under the approaches they employ in their components (section \ref{components}) or other model characteristics. This will help in developing a high-level understanding of the CMC models. A model-wise summary of the CMC models is provided in Appendix  \ref{appdx:model_details}.
\subsection{Trends in History Selection}
\label{attention_trends}
         Almost all of the current CMC models select conversational history based on a heuristic of considering $k$ immediate turns, often decided by performance such as BiDAF++ \cite{quac,dataset_comparison}, SDNet \cite{sdnet}, BiDAF++ w/ 2-ctx \cite{BERT_w/2-ctx} use last two turns as including the third turn degrades performance. History Attention Mechanism (HAM) based model \newcite{ham} uses a dynamic history selection policy by attending over contextualized representations of all the previous history turns at word-level or sequence-level and combining with current turn's representation as shown in Fig.~\ref{fig:ham}.
        
\subsection{Trends in History Modeling}
\label{history_modelling}
       How conversational history is integrated or used in the encoding process of contextual input embeddings can be used to classify CMC models. Different trends observed in this respect are described below. Some models may use a combination of these approaches.
       \begin{enumerate}
       \itemsep0em
           \item \textbf{Appending selected history questions and/or answers} (in raw form or text span indices) \textbf{to the current question} before encoding. QA tokens across turns should be distinguishable or separated when appending. Models DrQA+PGNet \cite{Coqa}, SDNet \cite{sdnet} and  RoBERTa + AT + KD \cite{Roberta+KD} append all history QA pairs separated by tokens like symbols $[Q]$ or $[A]$ such that new  $ Q_k^* = \{[Q], Q_1, [A], A_1, ...,  [Q], Q_{k-1}, [A],A_{k-1}, [Q], Q_k\}$. On the other hand, QuAC baseline model BiDAF++ w/ 2-ctx \cite{BERT_w/2-ctx} and GraphFlow \cite{GraphFlow} append only the history questions to the current question and encode relative dialog-turn number within each question embedding to differentiate. \newcite{quac} validate that this dialog-turn encoding strategy performs better in practice. 
        
            \item \textbf{Encoding context tokens with history answer marker embeddings (HAE)} before passing on for reasoning. These embeddings indicate if the context token is present in any conversational history answer or not, such as in BiDAF++ w/ 2-ctx \cite{quac}, GraphFlow \cite{GraphFlow}, BERT+HAE \cite{HAE} and HAM \cite{ham}. HAM encodes a dialog-turn encoded variant of HAE called \textit{Positional HAE}. It maintains a lookup table of history embeddings for every relative position from the current conversation and embeds the corresponding embedding if the token is found in that history answer, e.g. for the current question $q_k$ if a token is found in history answer $a_{k-2}$ then Positional HAE embedding at index 2 is encoded, otherwise embedding at index 0 is encoded. This setting is illustrated in Fig.~\ref{fig:bert_reasoning}.
            
            \item \textbf{Integrating intermediate representations generated in the reasoning modules} of selected history conversation turns to grasp the deep latent semantics of the history, rather than acting on raw inputs. This approach is also called the \textit{FLOW based approach}. The models that follow this approach are FlowQA \cite{flowQA}, FlowDelta \cite{flowDelta}, and GraphFlow \cite{GraphFlow}. GraphFlow encodes conversational histories into context graphs which are used by the reasoning module for contextual analysis. 
          \end{enumerate}
  
For contextual encoding, most of the models utilize one of the two types of encoders: (a.) Bidirectional sequential language models such as BiDAF \cite{BiDAF} or ELMo \cite{Elmo} (b.) Deep bidirectional transformer-based models such as BERT \cite{bert} or RoBERTa \cite{liu2019roberta}.

\subsection{Trends in Contextual Reasoning}
    \label{contextual_reasoning}
    While every CMC model has its unique flavor in integrating encoded representations of the query, history, and text contextually, some recurrent themes in reasoning can still be drawn. It is important to note that some of these themes will reflect state-of-the-art techniques around their release, which may now be obsolete. However, having their knowledge would prevent the re-exploration of those ideas. Following are the commonly observed themes:
    
    \vspace{0.5em}
    \noindent
    \textbf{A. Attention-based Reasoning with Sequence Models}
    \vspace{0.2em} 
    
    \noindent
    This was a common theme across MRC models until transformers \cite{transformers} were introduced and got rid of sequence modeling. Consequently, initial baseline models were based on this approach. 
    
    \noindent
    \textbf{CoQA baseline} \cite{Coqa} first involves DrQA \cite{drQA}, which performs BiLSTM based contextual integration over encoded tokens for extractive span, and later PGNet, that uses attention-based neural machine translation \cite{NMT} for abstractive answer reasoning. 
    \newline\noindent
    \textbf{QuAC baseline} \cite{quac} combines self-attention with BiDAF \cite{BiDAF} that performs reasoning via multi-layered bidirectional attention followed by multi-layered BiLSTM (BiDAF++).
    \newline\noindent
    \textbf{SDNet} \cite{sdnet} applies both inter-attention and self-attention in multiple layers,  interleaved with BiLSTM, to comprehend conversation context.
                 
    \vspace{0.5em}
    \noindent
    \textbf{B. FLOW based approaches}
    \vspace{0.2em} 
    
    \noindent
    Analogous to recurrent models which propagate contextual information through the sequence, FLOW is a sequence of latent representations that propagates reasoning in direction of the dialog progression by feeding intermediate latent representations, generated during reasoning in previous conversations, into contextual reasoning for the current question. This helps to leverage the reasoning effort of previous conversations as compared to using shallow history, such as directly appending history question-answers, where important contextual information in conversations may be lost due to the overwhelming input. There are \textit{two} major flow-based approaches based on the propagated latent representation.
    \begin{figure}[h!]
        \begin{subfigure}{0.48\textwidth}
        \centering
            \makebox[\textwidth]{\includegraphics[scale=0.032]{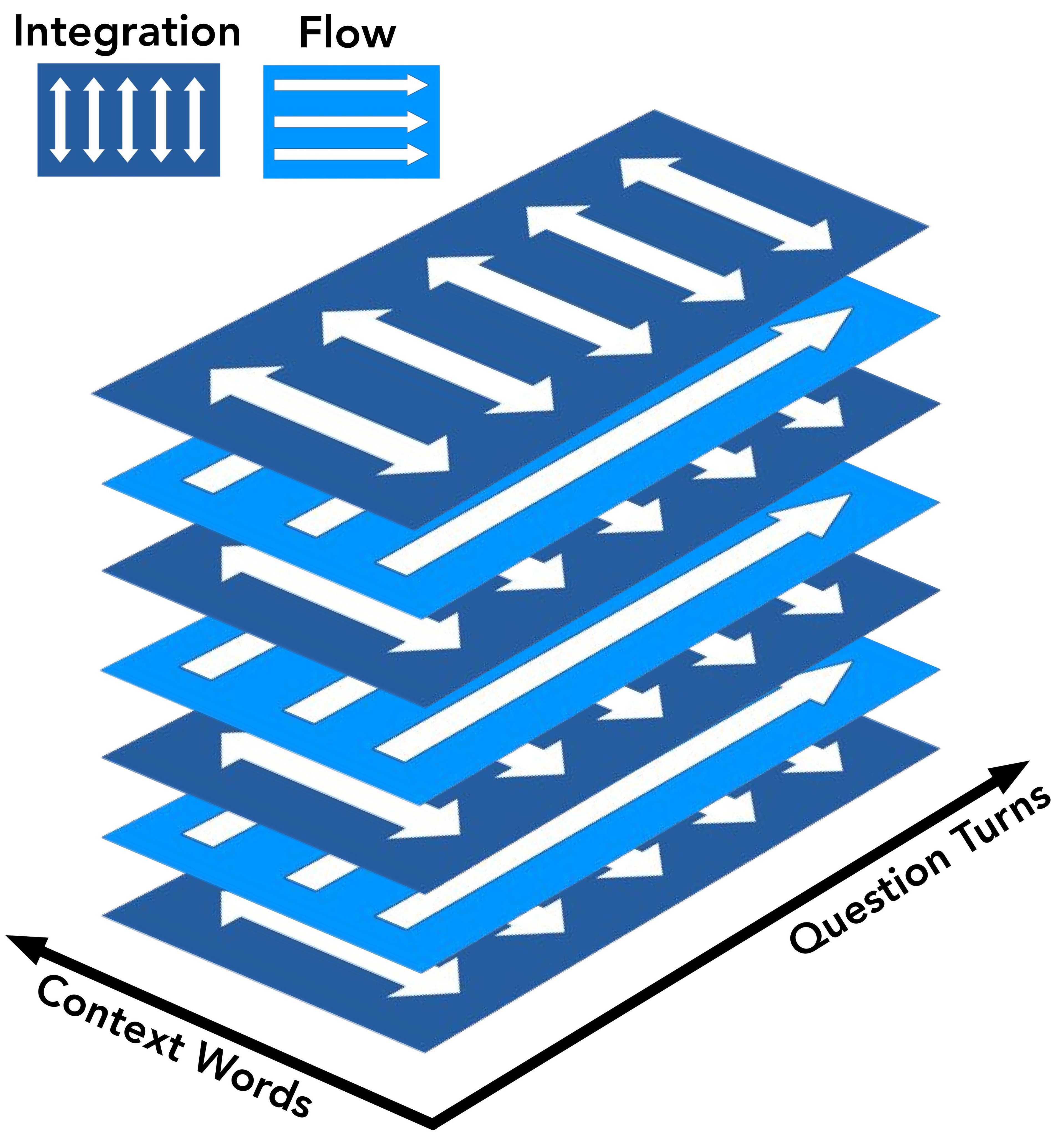}}
            \caption{Integration-Flow reasoning involves alternating computation between context integration (RNN over context) and FLOW (RNN over question turns).}
            \label{fig:IF}
        \end{subfigure}%
        \hspace{1em}
        \begin{subfigure}{0.48\textwidth}
        \centering
            \makebox[\textwidth]{\includegraphics[scale=0.13]{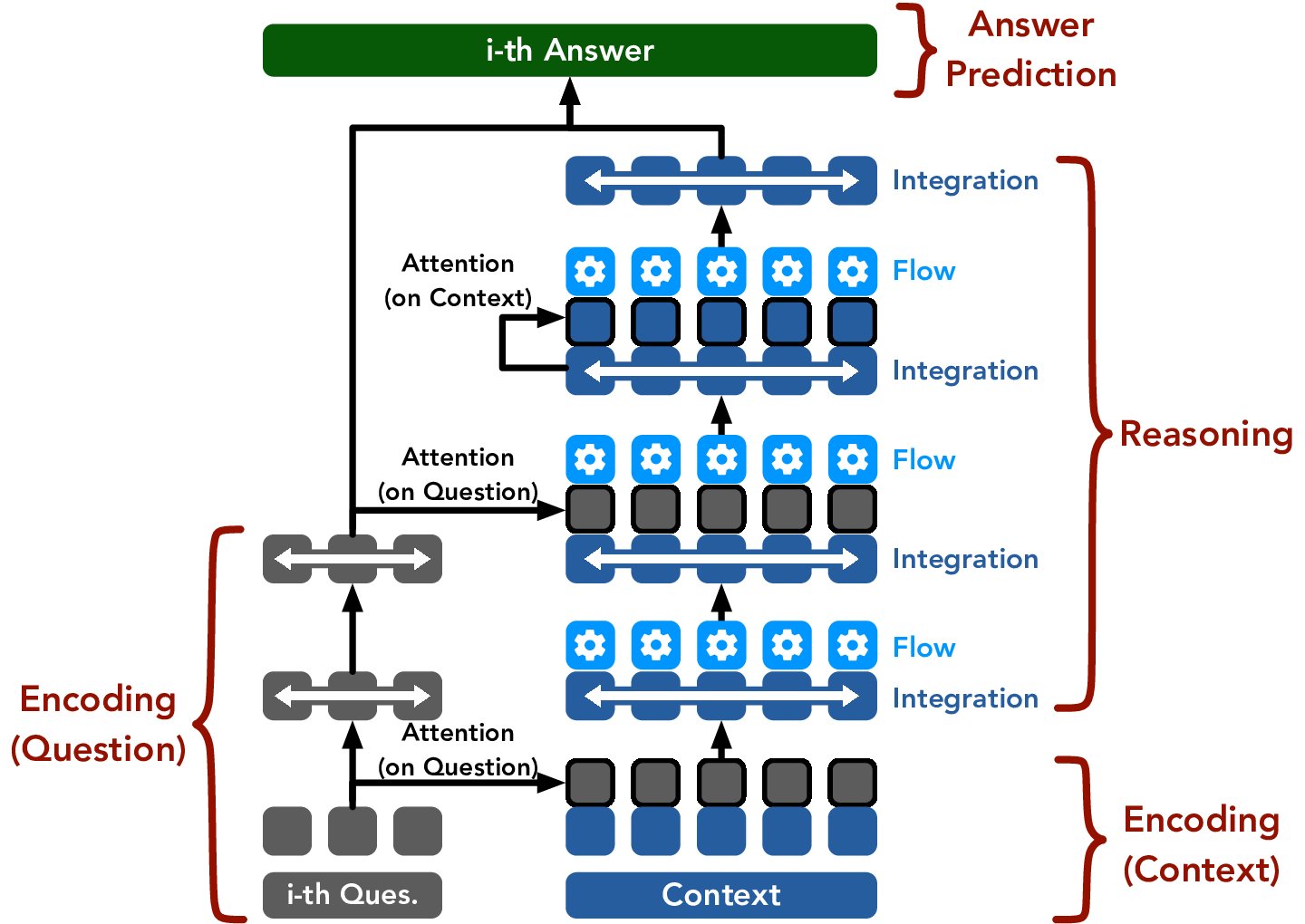}}
            \caption{FlowQA Architecture: Integration-Flow layers are alternated using cross-attention between the context and the question. Answer is predicted on the final concatenated output.}
            \label{fig:flowQA}
        \end{subfigure}
        \caption{Illustration of the Integration-Flow based reasoning in FlowQA. Source : \cite{flowQA} }
    \end{figure}
    \begin{enumerate}
    \itemsep0em
        \item \textbf{Integration-Flow (IF)}: This mechanism uses contextualized embeddings as the propagated latent representation. FlowQA \cite{flowQA} which also introduced the idea of FLOW, involves sequential processing along context tokens in parallel to the question turns followed by sequential processing in direction of the question turns (Flow), in parallel to context tokens as illustrated in Fig.~\ref{fig:IF}. FlowQA employs multiple IF layers interleaved with self and cross attentions to reason over encoded embeddings (Fig.~\ref{fig:flowQA}). Recently released FlowDelta \cite{flowDelta} is an improvement on the IF approach that uses the similar FlowQA architecture and achieves better results. Instead of passing the latent representation directly, as in FlowQA, FlowDelta passes the information gain (the difference between the latent representation of previous 2 layers) with the intuition that information gain would allow the model to focus on more informative cues in context.

        \item \textbf{Integration-GraphFlow (IG)}: GraphFlow \cite{GraphFlow} claims that the IF mechanism does not mimic human reasoning, as it first performs reasoning in parallel for each question, and then refines the reasoning results across different turns. Therefore, they use dynamically constructed, question-aware context graphs for each turn as the propagated latent representation. Processing through this flow (called GraphFlow) is facilitated by applying GNNs \cite{GNN} on the current context graph and previous context. To capture local interactions among consecutive words in context before feeding to a GNN, a BiLSTM is applied for contextual Integration. GraphFlow architecture alternates this mechanism with co-attention over the question and GNN output. This is illustrated in the figure provided in Appendix~\ref{appdx:graphflow}.
    \end{enumerate}
 
    \vspace{0.5em} \noindent
    \textbf{C.  Contextual Integration using Pre-trained Language Models}
    \vspace{0.2em} 
    
    \noindent
    Large-scale pre-trained LMs such as BERT \cite{bert}, GPT \cite{gpt} and RoBERTa \cite{liu2019roberta}, have become the current state-of-the-art approaches for contextual reasoning in CMC models, with leaderboards of both datasets stacked with these models or their variants. The approach is based on the fine-tune BERT-based MRC modeling outlined by \newcite{bert}, in which question and context are packed together (with marker embeddings to distinguish) in an input sequence to BERT that outputs contextualized question-aware embeddings for each input token. 
    
    Using pre-trained models for reasoning is advantageous in two aspects: \textit{Firstly}, it simplifies the architecture by fusing encoding and reasoning modules into a single module. \textit{Secondly}, it provides a ready-to-tune architecture that abstracts out complex contextual interactions between query and context while providing sufficient flexibility to control interactivity via \textit{augmentation of input embeddings} i.e. concatenation of special embeddings to input tokens that signal the model to incorporate a desirable characteristic in contextualization.  
    
\begin{figure}[h]
    \begin{subfigure}{0.48\textwidth}
        \centering
        \makebox[\textwidth]{\includegraphics[scale=0.28]{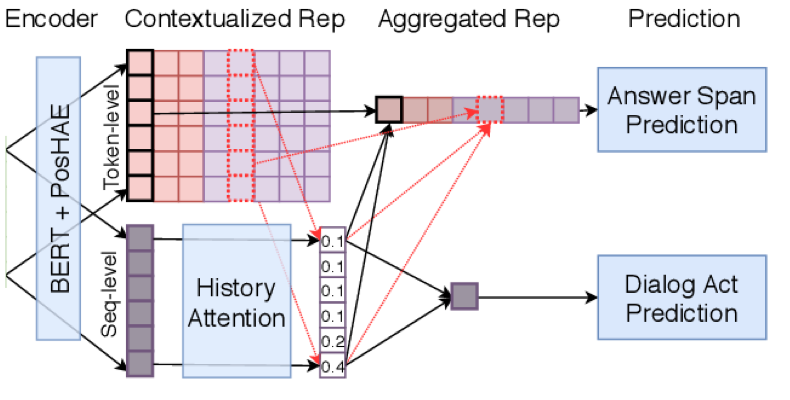}}
        \caption{HAM uses a dynamic attention-based history selection policy. Contextualized representations are generated by the model's encoder (BERT+PosHAE) for every history turn at word and sequence levels. Sequence-level embeddings are used to compute attention weights via scaled-dot product, and aggregate representations are generated by a weighted combination of embeddings of each turn in the proportion of their attention weights. Thus, attention weights help in determining the degree of selection (relevance) of each history turn.}
        \label{fig:ham}
    \end{subfigure}
    \hspace{1em}
    \begin{subfigure}{0.48\textwidth}
        \centering
        \makebox[\textwidth]{\includegraphics[scale=0.33]{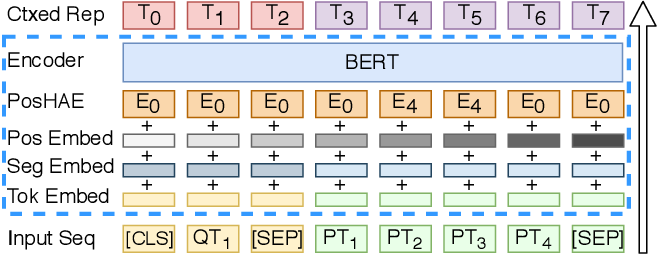}}
        \caption{ HAM's BERT based Encoder (Reasoning Architecture) for every conversation turn. The encoder is provided with input sequence consisting of query tokens (yellow) and context tokens (green) separated by \texttt{[SEP]}. It outputs contextualized representations $T_i$ corresponding to aligned question/passage tokens. The Token embeddings are augmented with segment embeddings(to differentiate query and context), positional embeddings (for distinct position in the sequence), and Positional HAE embeddings (for encoding history answer and relative conversational turn).}
    \label{fig:bert_reasoning}
    \end{subfigure}
    \caption{Illustration of (a) history selection module and (b) encoder/reasoning module of History Attention Mechanism (HAM) model \cite{ham}.}
\end{figure}

However, incorporating history into these models is a key challenge in this approach as most of the transformer models such as BERT only accept 2 segments ids in the input sequence. Based on recent research in CMC, two main trends in solving the history integration issue are discussed below:
\begin{enumerate}
\itemsep0em
    \item \textbf{Modify the input embeddings for a single-turn MRC model} to incorporate history. This is done by either appending the entire conversation to the question, such as \newcite{Roberta+KD} which uses RoBERTa \cite{RoBERTa} as the base model and truncates query if it exceeds the limit, or add special embeddings to highlight conversational history for the model, such as HAE \cite{HAE} embeds history answer embeddings with each context token if it is present in any of the history turns (detailed in section \ref{history_modelling}). This approach does not effectively use the model to capture interactions between every dialog-turn and context.
    
    \item  \textbf{Use separate model for each conversational turn} to capture one-to-one interaction between history and context, and merge the per-turn contextualized embeddings into aggregated history-aware embeddings. Two models follow this trend. \newcite{BERT_w/2-ctx} uses BERT models to capture contextual interaction for every question (history and current) and answer (2N+1 sequences for N turns) and concatenates all sequences together. Finally, it runs Bi-GRU \cite{bi-GRU} over the aggregated sequence to capture inter-turn interactions before sending for prediction. On the other hand, HAM \cite{ham} ignores the history questions and uses the current question as a query with positional History Answer Embeddings (section \ref{history_modelling}), thus generating one output sequence per conversation turn. Fig. \ref{fig:bert_reasoning} illustrates HAM encoder. The final sequence is generated using token-level soft-attention based aggregation across all per-turn contextualized sequences. 

\end{enumerate}
\subsection{Trends in Training Methodology} 
\label{training_meth}
Due to the multi-output nature of both CoQA and QuAC, multi-task training is quite common amongst CMC models, e.g. HAM \cite{ham} uses multi-task learning over QuAC to also predict dialog prediction and continuation acts, while GraphFlow \cite{GraphFlow} uses multi-task learning over CoQA to also predict question type. Besides, recently published \cite{Roberta+KD} achieved state-of-the-art results using RoBERTa, by applying multiple training techniques over CoQA. These consist of rationale tagging multi-task learning (predict if the token exists in CoQA evidence), Adversarial Training \cite{adversarialTraining}, and Knowledge Distillation \cite{KnowledgeDistillation}.

\section{Discussion}
\label{discussion}

\textbf{How does the research progress in CMC, a constrained setup, benefit the more into-the-wild domain of Conversational Search?} 
As stated by \newcite{HAE}, Conversational QA (and CMC) is a simplified setting of Conversational Search (ConvSearch), an information-seeking, “System Ask, User Respond” paradigm \cite{convSearch}, that does not focus on asking proactively. CMC, specifically, tries to address the challenges of NLU, via contextual encoding, reasoning, and handling conversational history, via history selection and modeling. In that aspect, CMC is a concrete enough setting for IR researchers to understand the change of information needs and interactivity between conversational cycles. 

\vspace{0.3em} \noindent
\textbf{Could Commonsense Reasoning improve CMC?} 
Commonsense Reasoning (CR) is based on the set of background information or world knowledge that an individual is intended to know or assume, and may be missing from context. On the other hand, Pragmatic reasoning, which the current CMC models cater to, is based on the derivation of explicit and implicit meanings within the context. The current MRC systems are nearing human performance on most datasets, however, they still perform poorly on single-turn CR based questions \cite{Record_dataset}. While there is recently increasing interest in CR in the single-turn MRC setting \cite{CosmosQM,mcscript,heterogeneous-commonsense-reasoning}, CMC remains relatively untouched. This may probably be due to the lack of foreknowledge requiring unanswerable questions (e.g. in SQuAD 2.0 \cite{squad2.0}) in current CMC datasets \cite{dataset_comparison}, suggesting a need for more complex CMC datasets that incorporate CR. However, humans annotators may often apply common-sense reasoning involuntarily while answering questions or comprehending, thus leaving room for incorporating CR in models. There seems to be no recent work that invalidates, experimentally, the role of CR in CMC. QuAC, for example, is drawn from articles on personalities, and current models still lag behind the human benchmark. It may be worth experimenting if adding domain knowledge or attributes about the context, like location and gender, help improve answering these questions.

\vspace{0.3em} \noindent
\textbf{Why did the paper focus on common trends across each component rather than a single overarching classification of CMC models?} The study of common trends in modeling, rather than a single overarching classification, helped in providing a multi-faceted view of CMC that can generalize on future models, and identify possible open-ended research questions, such as (a) For history selection, HAM \cite{ham} has proved to be both effective and intuitive in selecting relevant history turns. The application of this history selection approach on previous techniques that considered immediate K turns could be experimented with. (b) As mentioned in training methodology (section \ref{training_meth}), RoBERTa-based CMC model \cite{Roberta+KD} that used knowledge distillation and adversarial training achieved state-of-the-art CoQA results \cite{Coqa}. This suggests that different training approach along with multi-task learning improves the performance of base models. These procedures could be experimented with more advanced models such as HAM \cite{ham} and FlowDelta \cite{flowDelta}.


\section{Conclusion}
In this paper, we provide a holistic overview of Conversational Machine Comprehension (CMC), which has seen a surge of research in recent years, owing to advancements in neural language modeling and the introduction of large-scale conversational datasets such as CoQA \cite{Coqa} and QuAC \cite{quac}. We discuss the challenges that make CMC different from machine reading comprehension (MRC) and compare the multi-turn conversational datasets: CoQA and QuAC, based on different CMC characteristics. To develop a high-level understanding of all the existing approaches to tackle CMC, we synthesize a general model framework and analyze the common trends across all the published models, loosely based on the components outlined in the framework. Finally, we discuss some open questions that emerged during our research and which, in our view, can be explored further. This review could serve as a compendium for researchers in this domain and help streamline research in CMC.


\bibliographystyle{coling}
\bibliography{coling2020}

\newpage
\appendix 
\section{General statistics for CoQA and QuAC}
\label{appdx:coqa_quac_example}
\begin{figure}[h!]
    \begin{subfigure}{0.43\textwidth}
     \centering
        \makebox[\textwidth]{\includegraphics[scale=0.23]{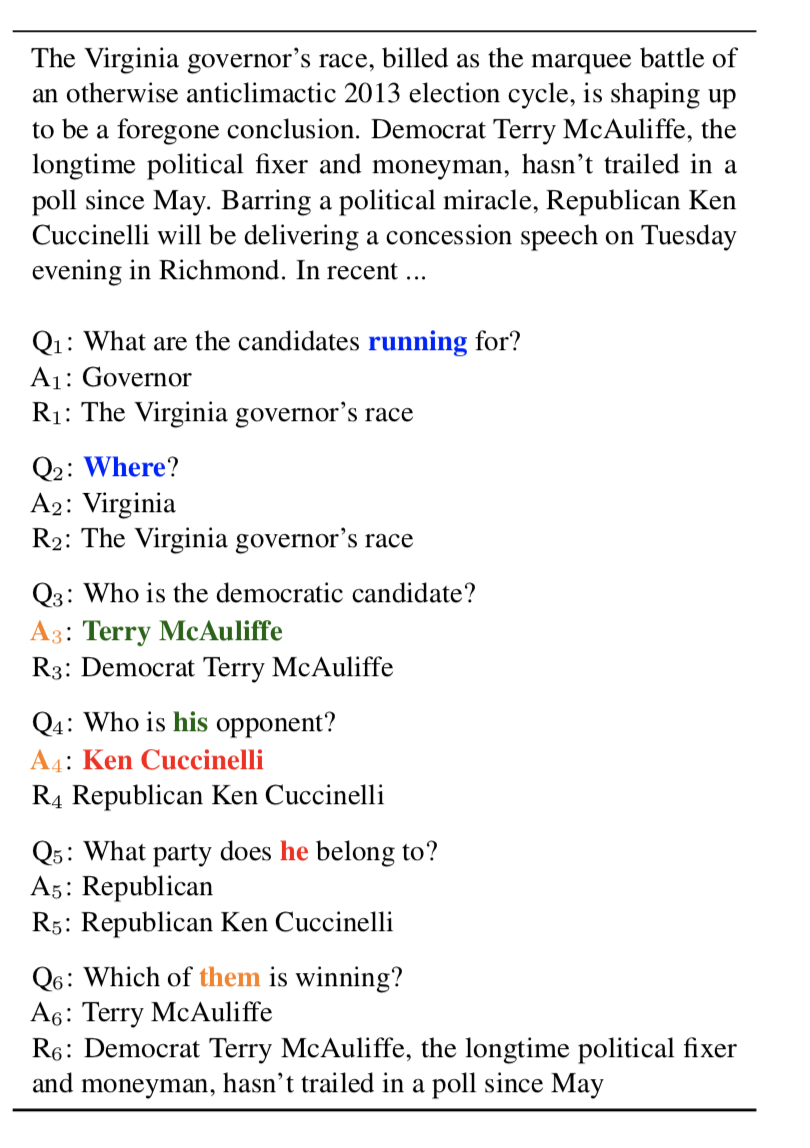}}
        \caption{A QA dialog example in the CoQA dataset. Every dialog is based on a context and each turn of the dialog contains a question ($Q_i$), an answer ($A_i$) and a rationale
($R_i$) that supports the answer. There is sufficient co-referencing between dialog turns as seen in this example -- \textit{`Where'} in $Q_2$ follows on the candidature mentioned in $Q_1$, \textit{`his'} in $Q_4$ points to $A_3$, \textit{`he'} in $Q_5$ references $A_4$, and \textit{`them'} in $Q_6$ refers to people mentioned in both $A_3$ and $A_4$. Source: \cite{Coqa}}
    \label{fig:coqa}
    \end{subfigure}%
    \hspace{1em}
    \begin{subfigure}{0.53\textwidth}
     \centering
    \makebox[\textwidth]{\includegraphics[scale=0.25]{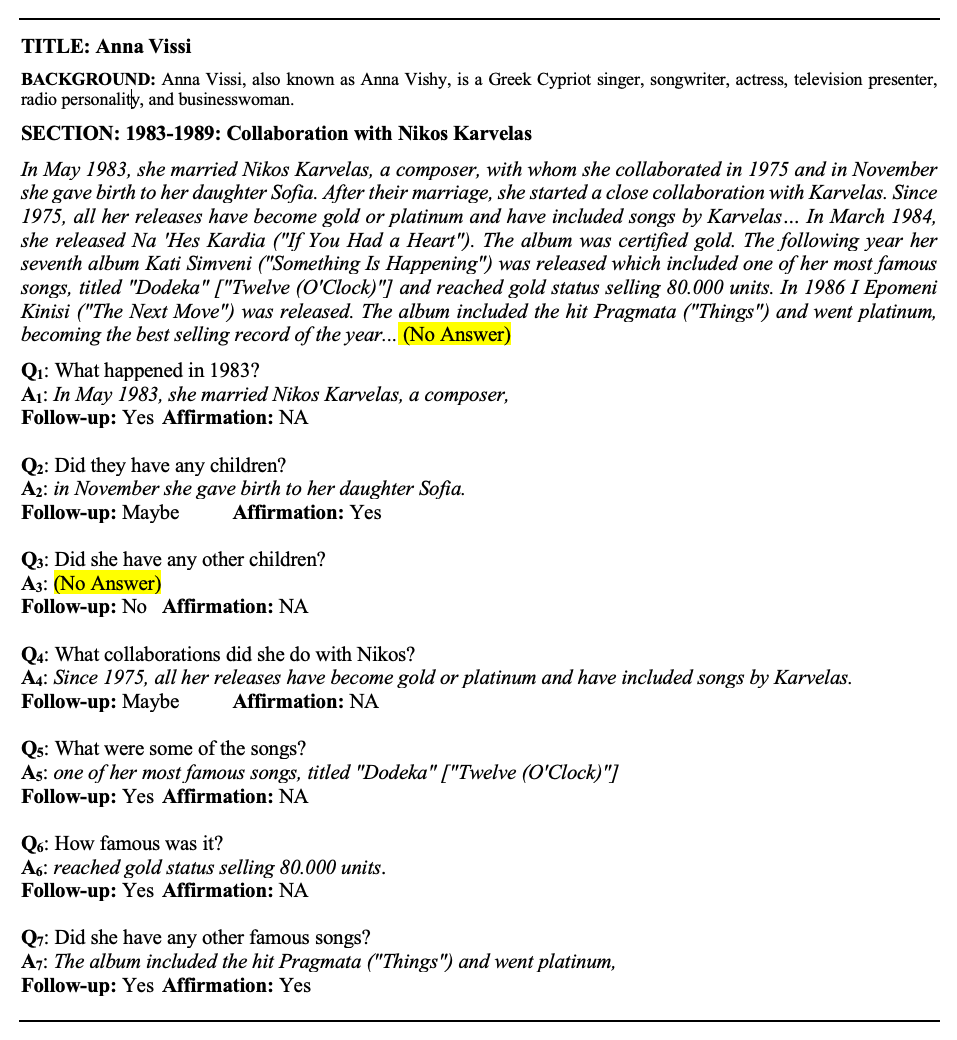}}
  \caption{A QA dialogue example from the QuAC dataset. Dialogs are prepared in a student-teacher setting over a section of a Wikipedia article, where the student questions based on the title, background, and start of the context and the teacher responds to the questions in the form of text spans from the context and dialogue acts. Dialog acts include follow-up i.e. whether the questioner should, could or should not cross-question, and, affirmation i.e. if the question can be answered as Yes/ No, or NA if it's not a yes/no kind of question. A `No Answer' token is appended to the context which is expected to be selected by the model in the case of unanswerable questions. The questions are open-ended due to the asymmetric nature of dataset. There is also sufficient co-referencing -- `she' in $Q_3$ refers to the protagonist and is a succession of $Q_2$, similarly $Q_7$ is a follow-up on $Q_5$, `it' in $Q_6$ refers to song mentioned in $A_5$.  Source: \cite{quac}}
    \label{fig:quac}
    \end{subfigure}%
    \caption{Illustrative examples from the conversational datasets-- CoQA and QuAC. }
    \label{fig:example_cq_quac}
\end{figure}

The dataset statistics are provided in Table~\ref{tab:comparison_quac_coqa}. An example from both datasets is provided in Fig.~\ref{fig:example_cq_quac}

\begin{table}[h!]
\centering
\resizebox{\columnwidth}{!}{%
\begin{tabular}{|p{4cm}|p{6cm}|p{8cm}|}
\hline
\textbf{Characteristic (average)}                   & \textbf{CoQA} & \textbf{QuAC}                                                      \\ \hline
\textit{Dataset source} &
  Passages collected from 7 diverse domains e.g. children stories from MCTest, news articles from CNN, Wikipedia articles, etc. &
  Sections from Wikipedia articles filtered in the ``people" category associated with subcategories like culture, animal, geography, etc. \\ \hline
\textit{Conversation setting} &
  Questioner-Answerer setting where both have access to the entire context. &
  Teacher-Student setting where the teacher has access to the full context for answering, while the student has only the title and summary of the article. \\ \hline
\textit{Question type}                              & factoid.      & open-ended, highly contextual.                                     \\ \hline
\textit{Answer type} &
  free-form with an extractive rationale. &
  Extractive span which can be yes/no or `No Answer'. It also provides dialog acts. \\ \hline
\textit{Total number of dialogs}                    & 8K            & 14K                                                                \\ \hline
\textit{Total number of questions}                  & 126K          & 100K                                                               \\ \hline
\textit{Context, Question and Answer token lengths} & 271, 5.5, 2.7 & 401, 6.5, 14.6                                                     \\ \hline
\textit{Turns per dialog}                           & 15.2          & 7.2                                                                \\ \hline
\textit{Unanswerable questions}                     & Very low and often
erroneously marked.      & Significant quantity of type 'missing info'                        \\ \hline
\textit{Evaluation metrics}                         & F1 scores for in-domain, out-of-domain and overall           & F1, Human Equivalence Quotient (HEQ) scores at question and dialog levels. \\ \hline
\end{tabular}%
}
\caption{A comparison of the multi-turn conversational datasets- CoQA \cite{Coqa} and QuAC \cite{quac} based on different characteristics defined in their papers and the analysis paper by \newcite{dataset_comparison}.}
\label{tab:comparison_quac_coqa}
\end{table}
\section{GraphFlow}
\label{appdx:graphflow}
\begin{figure}[h!]
    \centering
    \makebox[\textwidth]{\includegraphics[scale=0.23]{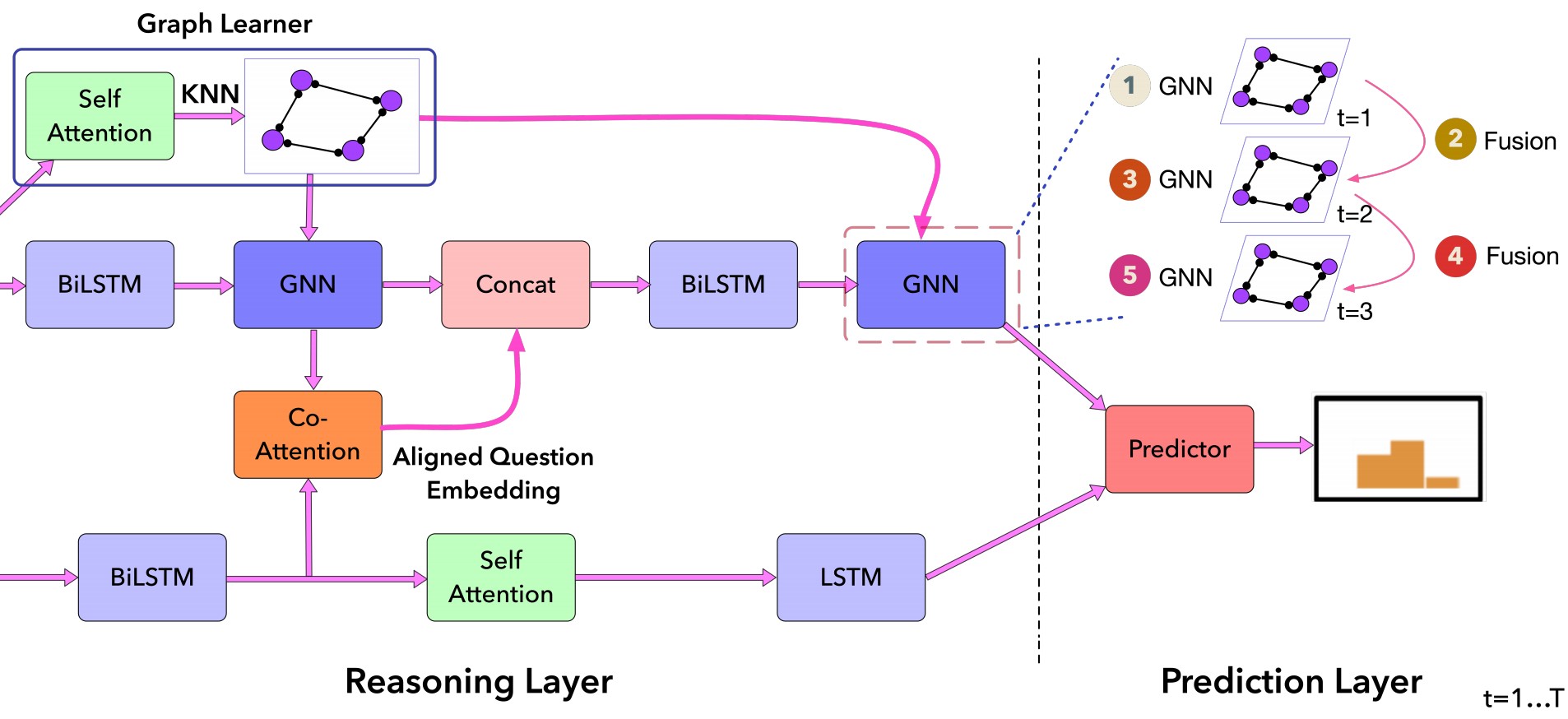}}
    \caption{Architecture of the Reasoning Layer of GraphFlow. Context graph-based flow sequence is processed using GNNs and alternated with bi-LSTM and co-attention mechanisms. Source : \cite{GraphFlow}}
    \label{fig:graphflow}
\end{figure}
\pagebreak
\section{A summary of the common CMC models}
\label{appdx:model_details}
The following tables provide a summary of the CMC models published on the CoQA\footref{coqa_l} and QuAC\footref{quac_l} leaderboards \footnote{Summarized as of July 1, 2020. Please note that the summary is only available for models that have an attached manuscript or repository. }. The table also provides a link to the official code repositories for the models. Although most of the models are published on both leaderboards, some models are very specific to one of the datasets. 

\begin{sidewaystable}[h!]
\centering
\begin{adjustbox}{center, max width=0.9\textwidth, max totalheight=\textheight}
\resizebox{\textwidth}{!}{%
\begin{tabular}{|M{4cm}|M{2cm}|M{2cm}|M{4cm}|M{8cm}|M{8cm}|M{5cm}|M{7cm}|}
\hline
\multirow{2}{*}{\textbf{Model}} &
  \multicolumn{2}{c|}{\textbf{Overall test F1 scores}} &
  \multicolumn{4}{c|}{\textbf{Approaches in}} &
  \multirow{2}{*}{\textbf{Repository}} \\ \cline{2-7}
 &
  \textbf{CoQA} &
  \textbf{QuAC} &
  \textbf{History Selection} &
  \textbf{History Modelling} &
  \textbf{Contextual Reasoning} &
  \textbf{Additional Training methods used (if any)} &
   \\ \hline
\textit{Human Performance} &
  88.8 &
  81.1 &
  N/A &
  N/A &
  N/A &
  N/A &
  N/A \\ \hline
\textit{RoBERTa + AT + KD \cite{Roberta+KD}} &
  90.7 (ensemble) 90.4 (single) &
  74.0 (ensemble) 73.5 (single) &
  All history Q\&A (until truncated) &
  Appending history Q\&A to the current question (truncate query if question tokens exceed) &
  Contextual Integration using single-turn, pre-trained LM - RoBERTa \cite{RoBERTa} &
  Rationale tagging multi-task (token prediction in CoQA evidence) along with Adversarial Training \cite{adversarialTraining} and Knowledge Distillation \cite{KnowledgeDistillation}. &
  No official code repo is given. \\ \hline
\textit{XLNet + Augmentation} &
  89.0 &
  71.2 &
  All history Q\&A (until truncated) &
  Appending history Q\&A to the current question (truncate query if question tokens exceed) &
  Contextual Integration using single-turn XLNet-large \cite{xlnet} (24L, 1024H, 16A) &
  None &
  \url{https://github.com/stevezheng23/xlnet_extension_tf} \\ \hline
\textit{BERT + Answer Verification} &
  82.8 &
  N/A &
  All history Q\&A (until truncated) &
  Appending history Q\&A to the current question &
  Contextual Integration using single-turn BERT-Base \cite{bert} (12L, 768H, 12A) &
  Additional answer verification multi-task as in \cite{answer-verification}. &
  \url{https://github.com/sogou/SMRCToolkit} \\ \hline
\textit{SDNet} &
  79.3 (ensemble) 76.6 (single) &
  N/A &
  Last 2 Q\&A &
  Appending history Q\&A to the current question &
  Attention-based Reasoning with sequential model- SDNet \cite{sdnet} &
  None &
  \url{https://github.com/Microsoft/SDNet} \\ \hline
\textit{BERT w/2-context \cite{BERT_w/2-ctx}} &
  78.7 &
  64.9 &
  Last 2 Q\&A &
  Append each Q (or A) to the current question for every BERT model. &
  Concatenates contextual outputs from BERT-Base models for every history question and answer (2N+1 sequences for N turns) and runs Bi-GRU over the aggregated sequence. &
  Answer type (Yes /No) prediction multi-task. &
  No official code repo is given. \\ \hline
\textit{HAM \cite{ham}} &
  N/A &
  65.4 &
  Dynamic selection policy by attending over all the previous history turns and deriving weights based on their contextual correlation with the current turn. &
  Concatenate context tokens with positional history answer marker embeddings corresponding to each dialog turn at a relative position from the current turn e.g. marker embedding 2 for turn 5 when the current turn is 7. Ignores history questions. &
  Aggregates contextual outputs from BERT-Base models for every answer (N sequences for N turns) using attention-based selection weights to obtain word-level and seq-level representations. &
  Dialog act (continuation and affirmation) prediction multi-task &
  \url{https://github.com/prdwb/attentive_history_selection} \\ \hline
\textit{Bert-FlowDelta  \cite{flowDelta}} &
  77.7 &
  65.5 &
  All history questions only &
  Integrate latent representations generated via contextual reasoning on the history turns (FLOW) &
  Each history turn is first passed through BERT whose last and second last layer outputs are input to separate FlowQA models. Reasoning is done using the Integration-Flow approach with information gain as the propagated FLOW representation. &
  None &
  \url{https://github.com/MiuLab/FlowDelta} \\ \hline
\textit{GraphFlow \cite{GraphFlow}} &
  77.3 &
  64.9 &
  All the history Q\&A &
  Uses a mix of all modeling techniques- encodes relative dialog turn numbers within each question embedding and appends all to the current question, then concatenates answer marker embeddings to context tokens and finally encodes the turn into N context graphs which are used for reasoning. &
  The encoded context graphs per turn are used as the propagated FLOW representation in the Integration-GraphFlow architecture. &
  Answer type (Yes /No) prediction multi-task. &
  \url{https://github.com/hugochan/GraphFlow} \\ \hline
\textit{FlowQA \cite{flowQA}} &
  75.0 &
  64.1 &
  All history questions only &
  Integrate latent representations generated via contextual reasoning on the history turns (FLOW) &
  Reasoning is done using the Integration-Flow approach with latent history turn representation directly as the propagated FLOW representation (instead of information gain). &
  Unanswerability multi-task. &
  \url{https://github.com/momohuang/FlowQA} \\ \hline
\textit{BERT+HAE \cite{HAE}} &
  N/A &
  62.4 &
  Last K history answers (max K=11) &
  Concatenate context tokens with history answer marker embeddings to identify if the token is present at any dialog turn. Ignores history questions. &
  Contextual integration using a single-turn BERT-base &
  None &
  \url{https://github.com/prdwb/bert_hae} \\ \hline
\textit{BiDAF++ w/2-Context (QuAC baseline) \cite{quac}} &
  67.8 &
  60.1 &
  Last 2 Q\&A &
  Encodes relative dialog turn numbers within each question embedding and appends all to the current question. Concatenates answer marker embeddings to context tokens. &
  Combines self-attention with BiDAF \cite{BiDAF} followed by multi-layered BiLSTM to obtain contextualized representation. &
  Dialog act multi-task in QuAC. &
  Use implementation on \url{http://allennlp.org} \\ \hline
\textit{DrQA+PGNet (CoQA baseline) \cite{Coqa}} &
  65.1 &
  N/A &
  All history Q\&A &
  Appending history Q\&A to the current question &
  BiLSTM based contextual integration over encoded tokens for extracting evidence (DrQA), followed by PGNet \cite{PointerN} which uses attention-based neural machine translation for abstractive answer reasoning (selecting relevant tokens). &
  None &
  \url{https://github.com/stanfordnlp/coqa-baselines} \\ \hline
\textit{Vanilla DrQA \cite{drQA}} &
  52.6 &
  N/A &
  All history Q\&A &
  Appending history Q\&A to the current question &
  BiLSTM based contextual integration over encoded tokens for extractive span. &
  None &
  \url{https://github.com/facebookresearch/DrQA} \\ \hline
\end{tabular}%
}
\end{adjustbox}
\caption{Model-wise summary of all the published CMC models for QuAC and CoQA.}
\label{tab:model-based-summary}
\end{sidewaystable}

\end{document}